\documentclass[preprint,12pt]{elsarticle}

\usepackage[english]{babel}
\usepackage{hyperref}
\usepackage[]{graphicx}
\usepackage{epstopdf}
\usepackage{longtable}
\usepackage{lscape}
\usepackage{multirow}
\usepackage{color}
\usepackage{array}
\usepackage{calc}
\usepackage{natbib}
\usepackage{amsfonts}
\usepackage{placeins}
\usepackage{epstopdf}
\usepackage[table]{xcolor}
\usepackage{arydshln}
\usepackage{eurosym}
\usepackage{float}
\usepackage{amssymb}
\usepackage{bm}
\usepackage{algorithm}
\usepackage{algorithmic}
\usepackage{mathptmx}
\usepackage{lineno,hyperref}
\usepackage{underscore}

\newcommand{\INDSTATE}[1][1]{\STATE\hspace{#1\algorithmicindent}}

\begin{document}

\begin{frontmatter}



\title{Learning Optimal Data Augmentation Policies via Bayesian Optimization for Image Classification Tasks}


\author[address1,address2]{Chunxu Zhang}
\ead{zhangcx2113@mails.jlu.edu.cn}
\author[address1,address2]{Jiaxu Cui}
\ead{jxcui16@mails.jlu.edu.cn}
\author[address1,address2]{Bo Yang\corref{correspondingauthor}}
\ead{ybo@jlu.edu.cn}

\cortext[correspondingauthor]{Corresponding author}

\address[address1]{College of computer science and technology, Jilin University, Changchun, 130012, China}
\address[address2]{Key Laboratory of Symbolic Computation and Knowledge Engineering of Ministry of Education, Jilin University, Changchun, 130012, China}

\begin{abstract}
In recent years, deep learning has achieved remarkable achievements in many fields, including computer vision, natural language processing, speech recognition and others. Adequate training data is the key to ensure the effectiveness of the deep models. However, obtaining valid data requires a lot of time and labor resources. Data augmentation (DA) is an effective alternative, which can generate new labeled data based on existing data using label-preserving transformations. Although we can benefit a lot from DA, designing appropriate DA policies requires a lot of expert experience and time consumption, and the evaluation of searching the optimal policies is costly. So we raise a new question in this paper: how to achieve automated data augmentation at as low cost as possible? We propose a method named BO-Aug for automating the process by finding the optimal DA policies using the Bayesian optimization approach. We validate the BO-Aug on three widely used image classification datasets, including CIFAR-10, CIFAR-100 and SVHN. Experimental results show that the proposed method can achieve state-of-the-art or near advanced classification accuracy at a relatively low policies search cost, and the searched policies based on a specific dataset are transferable across different neural network architectures or even different datasets. Code to reproduce our experiments is available at https://github.com/zhangxiaozao/BO-Aug.
\end{abstract}


\end{frontmatter}


\section{Introduction}\label{Introduction}
Deep neural networks are powerful machine learning systems, which have been demonstrated in a variety of tasks including image classification \cite{krizhevsky2012imagenet, he2015spatial}, object detection \cite{girshick2014rich, redmon2016you}, natural language processing \cite{collobert2008unified, collobert2011natural} and speech recognition \cite{hinton2012deep, graves2013speech} among others. Generally, the more data, the better deep neural networks can do. Insufficient amount of data can lead to model overfitting, which will reduce the generalization performance of the model on the test set. In recent years, many techniques \cite{hinton2012improving, srivastava2014dropout, bouthillier2015dropout, ioffe2015batch} have been proposed to help combatting overfitting such as dropout, batch normalization, L1/L2 regularization and layer normalization. Nonetheless, these techniques will also fall short when the data is limited, since the flexibility of the deep models is so high.

Labeling training examples by humans is a way to acquire a larger labeled dataset but difficult and time-consuming. Another way is data augmentation (DA), which refers to generate more labeled data based on the available samples using label-preserving transformations. DA allows us to train a deeper neural network which may alleviate overfitting and thereby improve the generalization performance of the model \cite{krizhevsky2012imagenet}. As its advantage, DA has been widely used in many domains, including natural language processing \cite{sennrich2015improving, hoang2018iterative, prabhumoye2018style}, speech recognition \cite{ko2015audio, rebai2017improving}, and especially computer visions applications \cite{krizhevsky2012imagenet, lecun2015deep, cirecsan2012multi}.

Despite we can benefit a lot from DA, how to design the optimal DA policies is still a challenging task. Low quality augmented samples produced by inappropriate DA policies may have the negative effect on the performance and robustness of the model. However, high quality DA policies require a wealth of expert experience and plenty of time. Moreover, the evaluation of searching the optimal DA policies is costly, and the policies found based on a specific model or dataset are usually not reusable, which will result in a need to re-search each time when DA policies are required, which further increases the cost of DA. In order to design efficient DA policies, some works \cite{tran2017bayesian, lemley2017smart, perez2017effectiveness, antoniou2017data, zhu2017data, ratner2017learning, cubuk2018autoaugment} have been proposed. Although the existing methods enable automated DA, the policies they generated are usually not reusable, or the cost of searching policies is high. Thus, we raise a new problem of how to automatically design DA policies at as low cost as possible.

\begin{figure}[!htb]
\centering
\includegraphics[width=10cm,height=5.5cm]{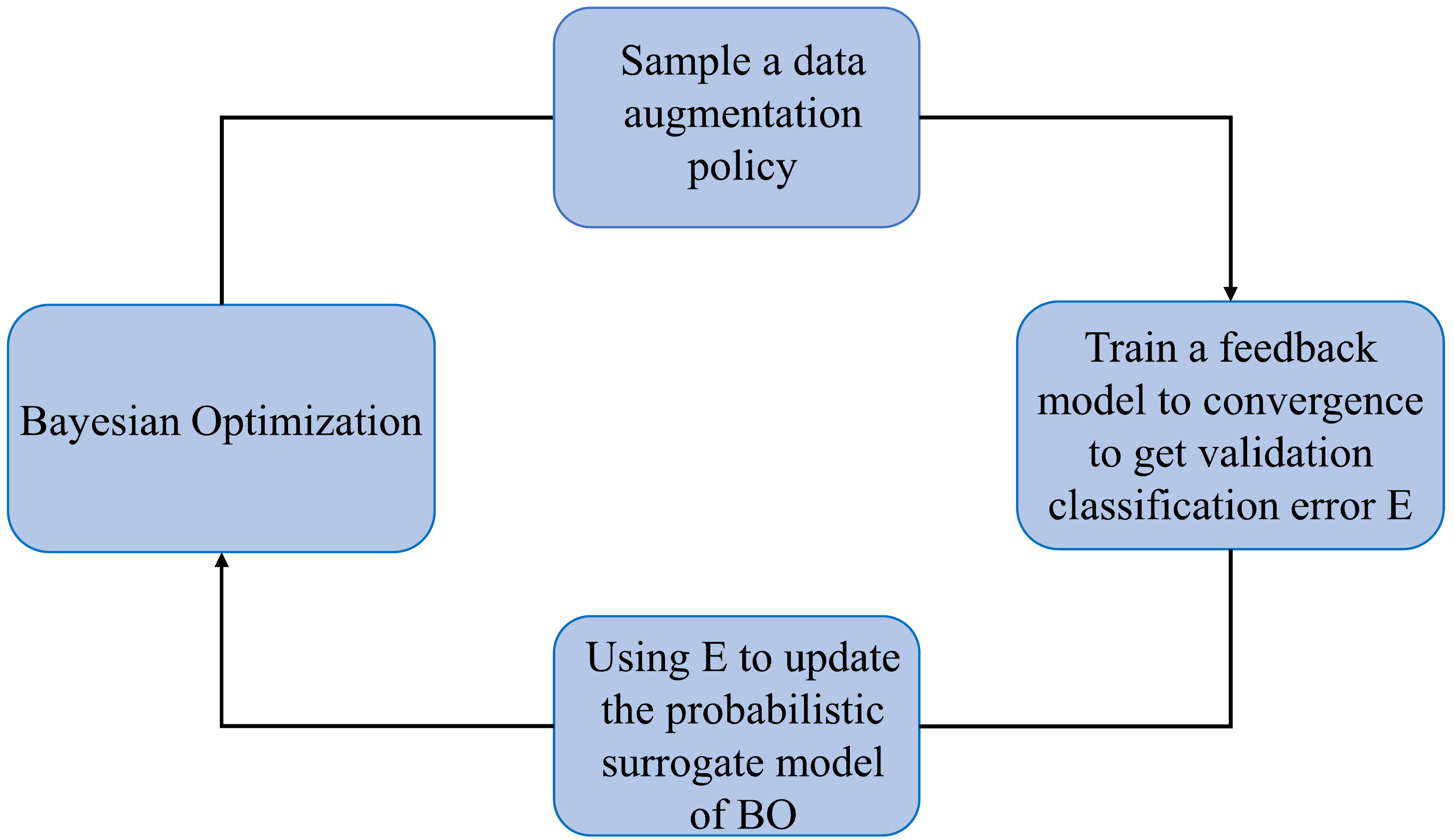}
\caption{Overview of the BO-Aug. Bayesian optimization can sample a promising DA policy in each iteration from policies search space. An image classification neural network feedback model, which uses the sampled policy, is trained to convergence on a validation dataset acquiring classification error E. Classification error E is used to update the probabilistic surrogate model of BO.}
\label{fig1}
\end{figure}

For tackling the above challenges, in this paper, we propose an automated DA method named BO-Aug in the context of image classification, where the learned policies can be easily used by different end discriminative models. Specifically, the method consists of two main components, policies search space design and search algorithm design. The search space contains lots of DA policies and each of them consists of 3 sub-policies, and each sub-policy is composed of 2 consecutive image transformation operations (e.g., shear, translation or rotation) and corresponding parameters (i.e., the probability and magnitude of applying the operation). The search algorithm adopted here is Bayesian optimization (BO) \cite{shahriari2016taking}, which is an effective global optimization algorithm. On each dataset, BO-Aug can find the optimal policies at a relatively low cost, which only needs 800 real evaluations, far less than the 15,000 real evaluations required by the current optimal automated DA algorithm \cite{cubuk2018autoaugment}. And the searched optimal policies can achieve state-of-the-art or near advanced classification accuracy. The overview of the BO-Aug is shown in Figure \ref{fig1}.

The main contributions of the proposed method can be summarized as:
\begin{itemize}
\item We propose an automated data augmentation method named BO-Aug which can automatically search the optimal DA policies on image classification tasks.
\item BO-Aug can find the optimal DA policies at a relatively low search cost, and the selected policies can achieve state-of-the-art or near advanced classification accuracy on several widely used image classification datasets, including CIFAR-10, CIFAR-100 and SVHN.
\item The found DA policies based on a specific dataset can be well transferred across different neural network architectures or even different datasets.
\end{itemize}

The paper is specifically organized as follows. In Section \ref{Background and Related Work}, we review the related work on data augmentation and Bayesian optimization. In Section \ref{BO-Aug: The proposed method} we propose our approach. Sections \ref{Experiments} gives the detailed experimental setup and results. Section \ref{Conclusion and Future Work} concludes the paper and discusses the future work.

\section{Background and Related Work}\label{Background and Related Work}
\subsection{Data Augmentation}\label{Data Augmentation}
For image classification tasks, insufficient amount of training data will result in model overfitting, thereby the trained model may not generalize well on validation dataset or test dataset. Though several regularization techniques \cite{hinton2012improving, srivastava2014dropout, bouthillier2015dropout, ioffe2015batch} have been proposed to combat model overfitting, they will still fall short when the data is limited. Data augmentation is another effective technique to relieve model overfitting, which generates new data based on existing dataset by label-preserve transformations to enlarge the training dataset. It has been an indispensable component of all recent large-scale image classification models.

Considering that manually designing DA transformations is time-consuming and labor-intensive, and inappropriate transformations may reduce the generalization performance of the model, some automated DA techniques have been proposed to alleviate this problem. Tran et.al \cite{tran2017bayesian} utilized the Bayesian method to generate new annotated training samples depending on the distribution learned from the training set. Lemley et.al \cite{lemley2017smart} proposed to create a network which focuses on learning augmentations that take advantage of the mutual information within a class so that it can generate augmented data during the training process of a target network. Generative adversarial network (GAN) \cite{perez2017effectiveness, antoniou2017data, zhu2017data} is also a common method used for generating additional data. Ratner et.al \cite{ratner2017learning} first train a GAN to generate sequences that describe DA policies, so that the learned transformation model can be used to perform data augmentation for any end discriminative model. Cubuk et.al \cite{cubuk2018autoaugment} proposed the AutoAugment, which used the reinforcement learning as the search algorithm to find the best policies in a designed policy search space, which is the most relevant approach with BO-Aug.

Despite AutoAugment \cite{cubuk2018autoaugment} has achieved the most advanced results on several image classification datasets, the biggest problem with this method is that the cost of searching for the optimal policies is too high. The search algorithm needs to update its parameters with the classification error of the child model (an image classification neural network model) as feedback, and reinforcement learning require too much feedback data to achieve reasonable performance. For each dataset, the final policies given by AutoAugment \cite{cubuk2018autoaugment} are obtained after 15,000 real evaluations. In addition to high computational cost, AutoAugment \cite{cubuk2018autoaugment} treated the problem of finding the best DA policies as a discrete search task which may lead to sub-optimal solution. In order to break these two restrictions, this paper proposed to use Bayesian optimization as the search algorithm to solve the data augmentation problem. We define the problem as a search problem in a contiguous space, that is, for the probability and magnitude parameters of the transformation operations, our method can obtain any desirable value within the parameter range including high precision decimals. The experimental results show that only 800 real evaluations are needed to achieve similar or even better model classification accuracy with BO-Aug for each dataset.

\subsection{Bayesian Optimization}\label{Bayesian Optimization}
Bayesian optimization (BO) \cite{shahriari2016taking} is an effective global optimization algorithm which has been widely applied in designing problems. This method is quite suitable to solve complex optimization problems where the objective functions cannot be expressed, or are non-convex, multimodal and expensive to evaluate. By designing appropriate probabilistic surrogate model and acquisition function, the BO framework can obtain the near optimal solution with only a few times of real evaluations. As an effective means to optimize complex black box problems, BO has been applied in many fields \cite{jiaxu2017BOsurvey}, including game and material design \cite{khajah2016designing, frazier2016bayesian}, recommendation system \cite{li2010contextual, vanchinathan2014explore}, robotics, embedded systems and system design \cite{akrour2017local, torun2018global}, automatic algorithm configuration \cite{bergstra2011algorithms, snoek2012practical, swersky2013multi, mahendran2012adaptive} and so on.

BO is an iterative process consisting of three main steps: the first step is to select the next most potential evaluation point by maximizing the acquisition function; the second step is to evaluate the selected evaluation point using the real evaluation model; in the third step, the newly obtained input-observation value pair will be added to the historical observation set, and the probabilistic surrogate model is updated to prepare for the next iteration. See section \ref{Search Algorithm Design} for more details.

\section{BO-Aug: The proposed method}\label{BO-Aug: The proposed method}
We model the problem of searching the optimal data augmentation policies as a search problem in the continuous space. In our implementation, a policy is composed of 3 sub-policies and each sub-policy consists of 2 consecutive image transformation operations and corresponding parameters, the probability and magnitude of the applied operation. Our goal is to find the optimal DA policies at as low cost as possible. Essentially, what we want to optimize is a black box function that takes the data augmentation policy as input and the performance of the policy on the neural network classification model as output. This objective function is a multimodal function, so we aim to find those points that make the function values optimal with as few BO iterations (i.e., real policy evaluation number of times) as possible. The objective function expression is as follows,
\begin{equation}
    {p^*} = \mathop {\arg \min }\limits_{p \in P} f(m,u(p,d))
\end{equation}
Here, $p$ denotes the DA policy, $P$ denotes the policies search space, $m$ denotes an image classification model, $d$ denotes an image classification dataset, $u$ represents the process of applying a policy on a dataset to generate a new transformed dataset, and $f$ represents the process of training an image classification feedback model on a dataset and obtaining the classification error.

\subsection{Policies Search Space Design}\label{Policies Search Space Design}
\begin{figure}[!htb]
\centering
\includegraphics[width=10cm,height=8.5cm]{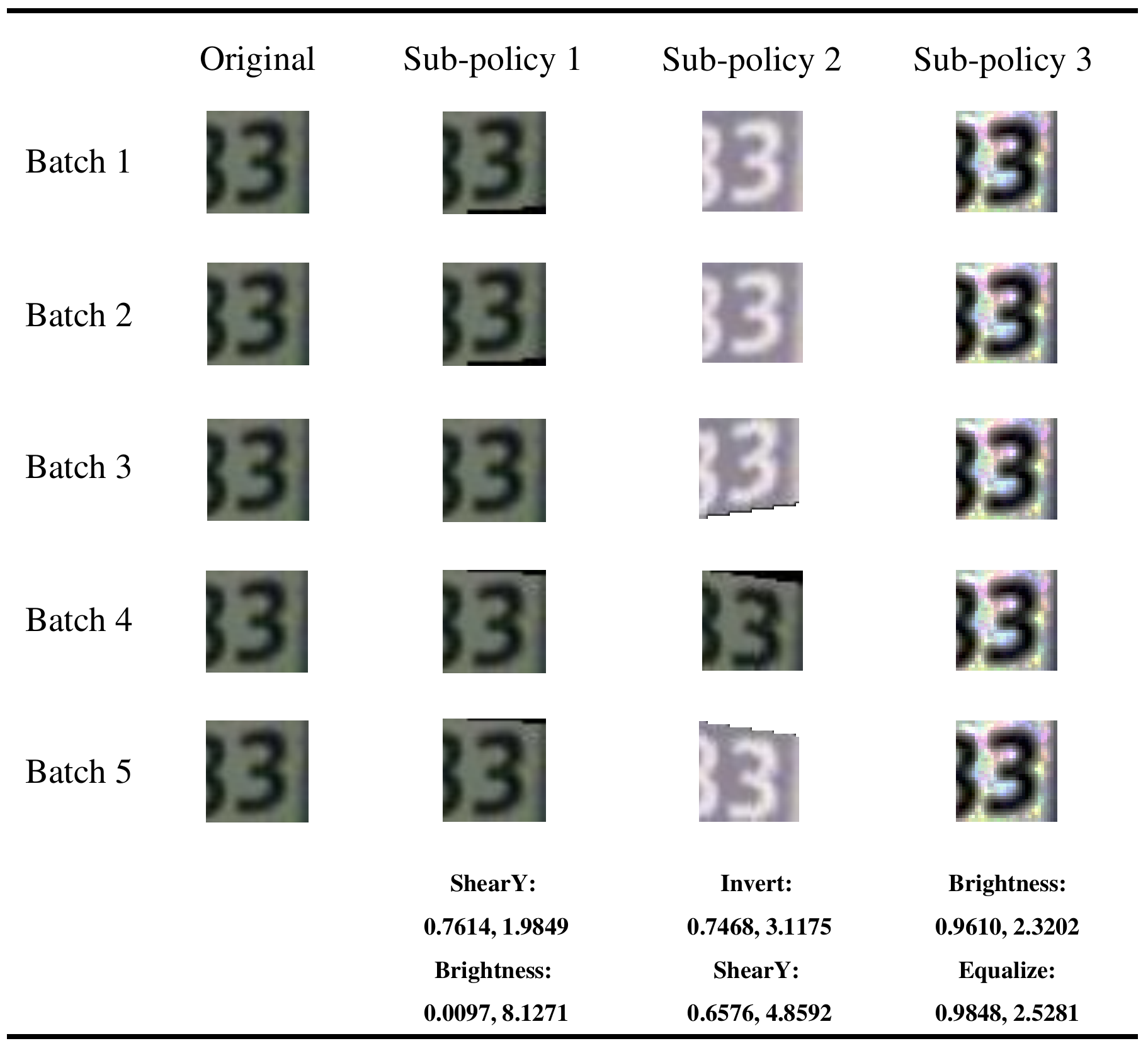}
\caption{One of the optimal policies found on SVHN\cite{netzer2011reading}. The learned policy can be used to generate augmented samples given an original image during the training process of deep models. The policy is composed of 3 sub-policies, and each sub-policy consists of 2 consecutive image transformation operations. For each operation, there are 2 corresponding parameters: the probability of applying the operation and the magnitude of the application. During every mini-batch training process, each image randomly selects a sub-policy from all available sub-policies with the same probability for data augmentation transformation. Note that one image can be transformed differently in different mini-batches even with the same sub-policy. In addition, the operation has an application probability so that it is not necessarily used even if an image have selected it. Both aspects have greatly increased the stochasticity of DA.}
\label{fig2}
\end{figure}

The image transformation operations used in this paper come from Python Imaging Library(PIL)$\footnote{https://pillow.readthedocs.io/en/5.1.x/}$, which is an image processing standard library that contains most of the basic image processing operations. Without loss of generality, we adopted all the 14 transformation functions in PIL. These functions accept an image and corresponding operation parameters as input, and output an transformed image. All the operations adopted here are shown in Table \ref{tab1}.
\begin{table}[!htb]
\footnotesize
\centering
\caption{\label{tab1}List of all image transformations that can be chosen during the search process. Additionally, the values of magnitude for each operation are shown in the third column. Some transformations do not use the magnitude information (e.g. Invert and Equalize).}
%\begin{ruledtabular}
\centering
\resizebox{\textwidth}{!}{
\begin{tabular}{llc}
%{\hsize}{@{}@{\extracolsep{\fill}}lp{10cm}lp{1cm}cp{0.01cm}}
\hline
\textbf{Operation Name}&\textbf{Description}&\textbf{Range of magnitudes}\\
\hline
ShearX(Y)&Shear the image along the horizontal(vertical) axis with rate&[-0.3, 0.3]\\
 &\emph{magnitude}.& \\
TranslateX(Y)&Translate the image in the horizontal(vertical) direction by \emph{mag-}&[-150, 150]\\
 &\emph{nitude} number of pixels.& \\
Rotate&Rotate the image \emph{magnitude} degrees.&[-30, 30]\\
AutoContrast&Maximize the image contrast,  by making the darkest pixel& \\
 &black and lightest pixel white.& \\
Invert&Invert the pixels of the image.& \\
Equalize&Equalize the image histogram.& \\
Solarize&Invert all pixels above a threshold value of \emph{magnitude}.&[0, 256]\\
Posterize&Reduce the number of bits for each pixel to \emph{magnitude} bits.&[4, 8]\\
Contrast&Control the contrast of the image. A \emph{magnitude=0} gives a gray&[0.1, 1.9]\\
 &image, whereas \emph{magnitude}=1 gives the original image.& \\
Color&Adjust the color balance of the image, in a manner similar to&[0.1, 1.9]\\
 &the controls on a colour TV set. A \emph{magnitude}=0 gives a black \&& \\
 &white image, whereas \emph{magnitude}=1 gives the original image.& \\
Brightness&Adjust the brightness of the image. A \emph{magnitude}=0 gives a black&[0.1, 1.9]\\
 &image, whereas \emph{magnitude}=1 gives the original image.& \\
Sharpness&Adjust the sharpness of the image. A \emph{magnitude}=0 gives a&[0.1, 1.9]\\
 &blurred image, whereas \emph{magnitude}=1 gives the original image.& \\
\hline
\end{tabular}}
\end{table}
For every operation, the execution probability is always between 0 and 1, and the range of operation magnitude depends on the specific operation type. For convenience, we set the value range of all operations¡¯ magnitude between 0 and 9 during policies search process, and then converted it to the value within the real magnitude range when using policies. There are several operations which do not contain magnitude information such as AutoContrast, Invert and Equalize. Since we treat the problem of finding the optimal DA policies as a continuous optimization problem, the operation's parameters can take any value within the parameters range including high precision decimals. This may help us to find more accurate parameters configuration that may beyond the handcrafted design. Figure \ref{fig2} illustrates an optimal DA policy found by the proposed method on SVHN dataset \cite{netzer2011reading}.

Considering that each sub-policy is composed of 2 consecutive operations, and there are 14 operations in total, so we can represent the operations in one sub-policy by a number between 0 and 196 (See Table \ref{tab9} for details.).
\begin{table}[!htb]
\footnotesize
\centering
\caption{\label{tab9} Examples of mapping relationship between the value of the first dimension of a sub-policy and corresponding 2 operation types.}
\centering
\begin{tabular}{lll}
\hline
\textbf{Opers}&\textbf{Operatio{n_1}}&\textbf{Operatio{n_2}}\\
\hline
0-1\qquad \qquad \qquad \qquad \qquad&ShearX\qquad \qquad \qquad \qquad \qquad&ShearX\\
1-2&ShearX&ShearY\\
2-3&ShearX&TranslateX\\
3-4&ShearX&TranslateY\\
4-5&ShearX&Rotate\\
5-6&ShearX&Solarize\\
6-7&ShearX&Posterize\\
7-8&ShearX&Contrast\\
8-9&ShearX&Color\\
9-10&ShearX&Brightness\\
10-11&ShearX&Sharpness\\
11-12&ShearX&AutoContrast\\
12-13&ShearX&Invert\\
13-14&ShearX&Equalize\\
\vdots&\vdots&\vdots\\
114-115&Color&TranslateX\\
\vdots&\vdots&\vdots\\
195-196&Equalize&Equalize\\
\hline
\end{tabular}
%\end{ruledtabular}
\end{table}
For each of the operation, there are two parameters associated with it. So a sub-policy can be represented by a 5-dimensional vector ${p_{sub} = [opers,pr{o_1},{m_1},pr{o_2},{m_2}]}$. The first dimension of the vector $opers$ represents the types of two operations of the sub-policy. The second dimension $pr{o_1}$ and the third dimension ${m_1}$ respectively represent the probability and magnitude of the first applied operation, and the fourth dimension $pr{o_2}$ and the fifth dimension ${m_2}$ represent the second operation's parameters jointly. Here is an example of using a 5-dimensional vector to represent a sub-policy (taken from a real sub-policy selected by BO): {[114.8650, 0.7610, 1.6081, 0.5414, 7.3520]} represents that the sub-policy consists of the Color and TranslateX two operations, the probability of the two operations are 0.7610 and 0.5414, respectively, and 1.6081 and 7.3520 are the magnitude parameters. Since a DA policy consists of 3 sub-policies, the single policy can be represented as a 15-dimentional vector $p = ({p_{sub - 1}},{p_{sub - 2}},{p_{sub - 3}})$. That is, policies search space is $P \subseteq {\Re ^{15}}$. Our goal, is to find 8 such policies totally in order to increase diversity.

\subsection{Search Algorithm Design}\label{Search Algorithm Design}
The search algorithm we used here is Bayesian optimization (BO). It starts by considering a set of known observations of the evaluation model. In the data augmentation scenario, these observations are the returned values of the real evaluation model on one policy. BO fits a probabilistic surrogate model to those observations, and then selects a new point to evaluate by maximizing the acquisition function over the parameter space. The selected point will be evaluated by the evaluation model and added to the existing observations set. This process will be executed iteratively until the predefined number of iterations is reached or the termination condition set by person is met. The complete iterative process and details of the search algorithm are shown in Algorithm \ref{algorithm}.
\begin{algorithm}[!htb]
%\algsetup{linenosize=\normalsize}
%\scriptsize
\caption{Searching for optimal DA policies}
\label{algorithm}
\begin{algorithmic}[1]
\REQUIRE ~~\\
$Initnum$: the number of initial known observations, here we take the value 10\\
$Iternum$: the number of iterations of the BO, here we take the value 90\\
\ENSURE ${P_{opt}}$, the final selected optimal DA policies\\
\STATE ${P_{opt}} = \{ \}$
\STATE \textbf{for} $T = 1,2,...,8$ \textbf{do}
\INDSTATE ${D_{init}} = \{ ({p_i},{y_i})|i = 1,...,Initnum\}$, where ${p_i} = RandomInit$ and ${y_i}=f(m,$\\
\quad $u({p_i},d))$ //$RandomInit$ is a method to initialize $p$ randomly
\INDSTATE ${D_0} = {D_{init}}$
\INDSTATE \textbf{for} ${\rm{t  =  1,2,}}...{\rm{,Iternum}}$ \textbf{do}
\INDSTATE[2] maximize EI to get the next evaluation point ${p_t}$
\INDSTATE[2] ${y_t} = f(m,u({p_t},d))$
\INDSTATE[2] ${D_t} = {D_{t - 1}} \cup ({p_t},{y_t})$
\INDSTATE[2] update GP
\INDSTATE \textbf{end for}
\INDSTATE ${P_{opt}} = {P_{opt}} \cup \{ {P_{opt - T}}\}$ // $P_{opt-T}$ is the optimal policies set found at the $Tth$\\
\quad time run of BO
\RETURN ${P_{opt}}$
\end{algorithmic}
\end{algorithm}
BO has several key elements: the real evaluation model, the choice of probabilistic surrogate model and acquisition function. Next, we will go through each of these components in detail.

In every iteration of BO, we need to execute a performance evaluation of the selected policy so that this feedback can be used to update the probabilistic surrogate model. Here we adopt the small WRN-40-2 (40 layers - widening factor of 2) \cite{zagoruyko2016wide} image classification model as the feedback model to evaluate that how good the selected policy is in improving the generalization of a model. The feedback model is trained with the augmented data generated by applying the selected policy on the training set. During the mini-batch training process, each image can randomly select one of the found sub-policies to apply with. After the training, the feedback model will be evaluated on the pre-reserved validation set to acquire the classification error as the feedback.

Probabilistic surrogate model is used to simulate unknown objective function, starting from hypothetical prior, by iteratively increasing the amount of information, correcting the prior, and thus obtaining a more accurate surrogate model. In this paper, we chose Gaussian process (GP) \cite{williams2006gaussian} as the surrogate model, a commonly used non-parametric model. Currently, GP has been widely used in regression, classification and many other domains where inferred black box functions are needed. A GP consists of a mean function and a semi-positive definite kernel function (or called covariance function). The kernel function is a measure of the ¡°similarity¡± between two points based on their locations in the parameter space of an objective function \cite{carr2016basc}. Here we used the Mat¨¦rn kernel function \cite{williams2006gaussian} that has high flexibility. And the priori mean function is usually assumed to be 0 for simplicity. It is worth nothing that we used the Markov chain Monte Carlo (MCMC) method to estimate the hyperparameters (i.e., scale parameter ${l_i}$ for each dimension of the BO's input) of GP \cite{williams2006gaussian}. Specifically, at each iteration of BO, we used MCMC to sample each hyperparameter multiple times and then respectively apply to the surrogate models, and the mean of these models will be used as the final surrogate model. MCMC takes into account the overall distribution of hyperparameters and usually achieves better results. We used expected improvement (EI) as the acquisition function, and found the point that maximized the EI function \cite{mockus1978application} with the help of the auxiliary optimizer covariance matrix adaptation evolution strategies (CMA-ES) \cite{hansen2001completely}. The EI function expression is shown in Equation 2.
\begin{equation}
{\alpha _t}\left( {x;{D_{1:t}}} \right) = \left\{ \begin{array}{l}
\left( {{v^*} - {\mu _{_t}}\left( x \right)} \right)\Phi \left( {\frac{{{v^*} - {\mu _{_t}}\left( x \right)}}{{{\sigma _t}\left( x \right)}}} \right) + {\sigma _t}\left( x \right)\phi \left( {\frac{{{v^*} - {\mu _{_t}}\left( x \right)}}{{{\sigma _t}\left( x \right)}}} \right),\sigma \left( x \right) > 0\\
0,\sigma \left( x \right) = 0
\end{array} \right.
\end{equation}
Here, ${v^*}$ is the current optimal function value, $\Phi ( \bullet )$ is the standard normal distribution cumulative density function, $\phi ( \bullet )$ is the standard normal distribution probability density function, ${\mu _t}(x)$ represents the posterior mean and ${\sigma _t}(x)$ represents the posterior standard deviation.

In our settings, BO can select one DA policy each time, and we run BO 8 times independently to find 24 sub-policies in total for the final training of the model on each dataset. We set the number of iterations for a single run of BO to 100, so we only need 800 real evaluations on each dataset, far less than 15,000 given by AutoAugment \cite{cubuk2018autoaugment}.

\section{Experiments}\label{Experiments}
In this section, we evaluate the BO-Aug on image classification tasks. We will answer the following 3 questions through the experiments:
\begin{enumerate}
\item What role does data augmentaion play in image classification tasks?
\item Can the proposed method outperform the baseline methods or obtain comparable performance in image classification tasks at a relatively low DA policies search cost?
\item Can the selected DA policies based on a specific dataset transfer well across different deep neural network models or even different datasets?
\end{enumerate}

\subsection{Dataset}\label{Dataset}
We validated our model performance based on three widely used image classification datasets which are CIFAR-10, CIFAR-100 and SVHN, respectively. The CIFAR-10 dataset \cite{krizhevsky2009learning} consists of 60,000 32x32 color images in 10 classes, with 6,000 images per class, and the training set contains 50,000 images and the remaining 10,000 images constitute the test set. The CIFAR-100 dataset \cite{krizhevsky2009learning} is similar to the CIFAR-10, except that it has 100 classes, each containing 600 images. The Street View House Numbers (SVHN) dataset \cite{netzer2011reading} contains 99,289 images which are all of small cropped digits whose size is 32x32. There are 73,257 images for training and 26,032 images for testing, and 531,131 additional images often used as the complementary training data. Following AutoAugment \cite{cubuk2018autoaugment}, in addition to the above mentioned three datasets, we also tested our model based on two reduced datasets denoted as Reduced CIFAR-10 and Reduced SVHN. Table \ref{tab2} presents
\begin{table}[!htb]
\footnotesize
\centering
\caption{\label{tab2} Summary statistics of datasets.}
\centering
\begin{tabular}{lccc}
\hline
\textbf{Dataset}&\textbf{Training set size}&\textbf{Test set size}&\textbf{Classes}\\
\hline
CIFAR-10&50,000&10,000&10\\
CIFAR-100&50,000&10,000&100\\
SVHN&604,388&26,032&10\\
Reduced CIFAR-10&4,000&10,000&10\\
Reduced SVHN&1,000&26,032&10\\
\hline
\end{tabular}
%\end{ruledtabular}
\end{table}
summary statistics for each of these datasets.

In order to save search time, we created a small dataset called reduced dataset taken from the original training dataset when experimenting on each dataset. For CIFAR-10, we randomly selected 4,000 images from the training set to form a Reduced CIFAR-10 dataset. The policies found on Reduced CIFAR-10 will be used on the CIFAR-10 and CIFAR-100 as well as on itself. For SVHN, we randomly selected 1,000 images from the training set to form a Reduced SVHN dataset in a similar way to Reduced CIFAR-10, and the selected policies will be used on Reduced SVHN and SVHN.

\subsection{Baseline Methods and Baseline Pre-process}\label{Baseline Methods and Baseline Pre-process}
We mainly considered Wide-ResNet \cite{zagoruyko2016wide} and Shake-Shake \cite{gastaldi2017shake} these two types of image classification models as the baseline models, which are currently the most advanced and widely used models on image classification tasks. Specifically, we selected the WRN-28-10, Shake-Shake (26 2x32d), Shake-Shake (26 2x96d) and Shake-Shake (26 2x112d) as the baseline classification models. In addition, we also choose Cutout \cite{devries2017improved} and AutoAugment \cite{cubuk2018autoaugment} as comparison methods. Specifically, Cutout refers to the baseline models adding Cutout preprocess operation during training. All the baseline methods are summarized in Table \ref{tab3}.
\begin{table}[!htb]
\footnotesize
\centering
\caption{\label{tab3} Summary of all the baseline methods.}
\centering
\resizebox{\textwidth}{!}{
\begin{tabular}{lc}
\hline
\textbf{Baseline Methods}&\textbf{Descriptions}\\
\hline
Wide-ResNet-28-10 \cite{zagoruyko2016wide}&\multirow{4}*{Baseline image classification models without DA policies}\\
Shake-Shake (26 2x32d) \cite{gastaldi2017shake}&\\
Shake-Shake (26 2x96d) \cite{gastaldi2017shake}&\\
Shake-Shake (26 2x112d) \cite{gastaldi2017shake}&\\
\hline
Cutout \cite{devries2017improved}&Baseline image classification models with Cutout data preprocess operation.\\
\hline
AutoAugment \cite{cubuk2018autoaugment}&Baseline image classification models with DA policies.\\
\hline
\end{tabular}}
%\end{ruledtabular}
\end{table}
For the baseline pre-process, we follow the procedure of state-of-the-art CIFAR-10 models: standardizing the data, zero-padding, random crops and using horizontal flips with 50\% probability. In addition to Reduced SVHN, we also add the Cutout \cite{devries2017improved} with 16x16 pixels on the rest datasets. For a single image, we first standardized it, then applied the learned DA policies, and finally used the other pre-processing operations described above.

\subsection{Experimental Environment and Hyperparameters configuration}\label{Experimental Environment and Hyperparameters configuration}
Our experiments are performed on Ubuntu16.04, a GeForce GTX 1080 Ti GPU, using TensorFlow (TF)$\footnote{https://www.tensorflow.org/versions/r1.9/api_docs/python/tf}$ deep learning framework. In the experiment, we found the optimal learning rate and weight decay hyperparameters which gave the best validation set accuracy and the remaining hyperparameters (i.e., the type of optimizer, the size of mini-batch and the training epochs) are consistent with those reported in the papers introducing the models \cite{zagoruyko2016wide, gastaldi2017shake}. The specific parameter configurations are shown in Table \ref{tab4}.
\begin{table}[!htb]
\footnotesize
\centering
\caption{\label{tab4} All hyperparameter configuration in the experiments. Due to our limited computing resources, we only spent a small amount of time to adjust the hyperparameters of the final image classification neural network, so the experimental results presented in this paper actually have room to further improvement.}
\centering
\begin{tabular}{llcc}
\hline
\textbf{Datasets}&\textbf{Model}&\textbf{Learning rate}&\textbf{Weight decay}\\
\hline
CIFAR-10&WRN-28-10&0.1&5e-4\\
CIFAR-10&Shake-Shake (26 2x32d)&0.01&1e-3\\
CIFAR-10&Shake-Shake (26 2x96d)&0.02&1e-3\\
CIFAR-10&Shake-Shake (26 2x112d)&0.01&1e-3\\
CIFAR-100&WRN-28-10&0.1&5e-4\\
CIFAR-100&Shake-Shake (26 2x96d)&0.02&1e-3\\
Reduced CIFAR-10&WRN-28-10&0.02&1e-3\\
Reduced CIFAR-10&Shake-Shake (26 2x96d)&0.02&1e-3\\
SVHN&WRN-28-10&0.01&3e-2\\
SVHN&Shake-Shake (26 2x96d)&0.02&5e-5\\
Reduced SVHN&WRN-28-10&0.1&1e-4\\
Reduced SVHN&Shake-Shake (26 2x96d)&0.01&1e-2\\
\hline
\end{tabular}
%\end{ruledtabular}
\end{table}

\subsection{Experimental Results}\label{Experimental Results}
\subsubsection{CIFAR-10 and CIFAR-100 Results}\label{CIFAR-10 and CIFAR-100 Results}
As mentioned above, for each dataset, we ran BO 8 times independently to find 24 sub-policies in total for the final models training. The policies found on Reduced CIFAR-10 are shown in Table \ref{tab5},
\begin{table}[!htb]
\footnotesize
\centering
\caption{\label{tab5} The optimal policies found on Reduced CIFAR-10. All the geometric-based operations are bolded.}
\centering
\begin{tabular}{lll}
\hline
 &\textbf{Operation 1}&\textbf{Operation 2}\\
\hline
Sub-policy 1&(Contrast, 0.6874, 3.5975)&(AutoContrast, 0.3478, 6.6923)\\
Sub-policy 2&(Brightness, 0.2875, 7.8932)&(AutoContrast, 0.9654, 4.9814)\\
Sub-policy 3&(Solarize, 0.9930, 3.0359)&\textbf{(Rotate, 0.6376, 3.0957)}\\
Sub-policy 4&(Posterize, 0.1359, 2.2361)&(Sharpness, 0.8544, 3.3977)\\
Sub-policy 5&\textbf{(TranslateY, 0.5520, 4.1526)}&(Brightness, 0.0143, 3.7034)\\
Sub-policy 6&\textbf{(ShearX, 0.9463, 8.9832)}&\textbf{(TranslateX, 0.5740, 1.9317)}\\
Sub-policy 7&(Equalize, 0.3051, 6.1926)&\textbf{(TranslateX, 0.6216, 0.0089)}\\
Sub-policy 8&\textbf{(TranslateY, 0.4527, 4.3223)}&(AutoContrast, 0.9646, 4.6787)\\
Sub-policy 9&\textbf{(ShearX, 0.8462, 2.6623)}&\textbf{(Invert, 0.5057, 2.5341)}\\
Sub-policy 10&(AutoContrast, 0.5957, 8.1685)&\textbf{(ShearY, 0.7342, 6.5833)}\\
Sub-policy 11&(Color, 0.0895, 0.7037)&(Equalize, 0.5103, 2.3937)\\
Sub-policy 12&\textbf{(Invert, 0.1966, 7.3815)}&(Solarize, 0.3622, 2.9833)\\
Sub-policy 13&\textbf{(ShearY, 0.5345, 5.0416)}&\textbf{(TranslateX, 0.9978, 7.4486)}\\
Sub-policy 14&\textbf{(Invert, 0.1813, 8.1266)}&(Posterize, 0.9963, 8.7576)\\
Sub-policy 15&(Posterize, 0.5928, 8.1990)&(Brightness, 0.9625, 5.8913)\\
Sub-policy 16&(Brightness, 0.5037, 3.2453)&(Brightness, 0.9853, 4.1967)\\
Sub-policy 17&\textbf{(TranslateY, 0.0772, 4.5035)}&\textbf{(TranslateY, 0.8807, 0.2262)}\\
Sub-policy 18&\textbf{(TranslateX, 0.2531, 4.3762)}&\textbf{(ShearX, 0.3437, 1.7062)}\\
Sub-policy 19&(AutoContrast, 0.7115, 4.3842)&(AutoContrast, 0.1873, 7.0988)\\
Sub-policy 20&(Equalize, 0.7708, 8.4010)&(Solarize, 0.2364, 1.3529)\\
Sub-policy 21&(Solarize, 0.7149, 5.3383)&(AutoContrast, 0.7863, 6.4704)\\
Sub-policy 22&(AutoContrast, 0.2122, 6.9200)&\textbf{(TranslateY, 0.0081, 6.5256)}\\
Sub-policy 23&(Equalize, 0.1361, 6.0541)&\textbf{(TranslateX, 0.9480, 3.6382)}\\
Sub-policy 24&\textbf{(TranslateX, 0.1695, 5.5734)}&(Brightness, 0.5921, 1.2058)\\
\hline
\end{tabular}
\end{table}
and they will be later used on CIFAR-10, CIFAR-100 and Reduced CIFAR-10.

Roughly speaking, the DA operations fall into two main categories: geometry-based operations and color-based operations. Similar to AutoAugment, the policies found on Reduced CIFAR-10 are most color-based operations (28 operations in total). Since the images in CIFAR-10 are very rich in color, and the main goal of the data augmentation is to help us increase the richness of the samples, the expansion from the color perspective is indeed straightforward. However, it is worth mentioning that the number of geometry-based operations is also close to half of the total (20/48). Intuitively, this also makes sense since the color difference of different categories is more obvious, using more geometry-based operations may help to increase the sample diversity within the category, which is beneficial to model classification. The transformation operation Invert was almost never used in all policies and operations AutoContrast, Brightness, sharpness, Equalize and Solarize were used frequently. Below, we will give the experimental results on CIFAR-10, CIFAR-100 and Reduced CIFAR-10 using policies found on Reduced CIFAR-10.

\noindent\textbf{CIFAR-10 Results}: The test set error on different neural network architectures is shown in Table 5, here we selected the WRN-28-10, Shake-Shake (26 2x32d), Shake-Shake (26 2x96d) and Shake-Shake (26 2x112d) as the evaluation models. As we can see, our method can achieve state-of-the-art classification accuracy on all models, but the policies used were obtained at a very low search cost.

\noindent\textbf{CIFAR-100 Results}: We also trained models on CIFAR-100 with the same policies found on Reduced CIFAR-10, the results are shown in table 5. Here we adopted the WRN-28-10 and the Shake-Shake (26 2x96) as the final evaluation models. Again, our method got pretty good results.

\noindent\textbf{Reduced CIFAR-10 Results}: Finally, we applied the same policies as above to train models on Reduced CIFAR-10, that is, to train the models on the 4,000 images used in policies selection. Following \cite{cubuk2018autoaugment}, the purpose of our experiment here is to compare with semi-supervised learning methods, but the biggest difference with them is that they use additional 46,000 unlabeled samples in addition to the 4,000 samples to train models jointly. The state-of-the-art error rate reported using semi-supervised method is 10.55\% \cite{miyato2018virtual}. Although AutoAugment achieved 10.04\% error rate using DA without additional unlabeled data, it has an extremely high computational overhead when choosing policies. In contrast, the evaluation cost of BO-Aug is just about one-twentieth of AutoAugment but achieved almost as good results. Detailed experimental results are shown in table \ref{tab6}.
\begin{table}[!htb]
\footnotesize
\centering
\caption{\label{tab6} Test set error rates(\%) on CIFAR-10, CIFAR-100 and Reduced CIFAR-10. Lower is better. All the baseline, Cutout and AutoAugment experimental results are reported in \cite{cubuk2018autoaugment}. Note that the policies used here are found on Reduced CIFAR-10.}
\resizebox{\textwidth}{!}{
\begin{tabular}{llcccc}
\hline
\textbf{Dataset}&\textbf{Model}&\textbf{Baseline}&\textbf{Cutout}&\textbf{AutoAugment}&\textbf{BO-Aug}\\
\hline
\textbf{CIFAR-10}&Wide-ResNet-28-10&3.87&3.08&2.68&\textbf{2.58}\\
&Shake-Shake (26 2x32d)&3.55&3.02&2.47&\textbf{2.43}\\
&Shake-Shake (26 2x96d)&2.86&2.56&1.99&\textbf{1.98}\\
&Shake-Shake (26 2x112d)&2.82&2.57&\textbf{1.89}&\textbf{1.89}\\
\textbf{CIFAR-100}&Wide-ResNet-28-10&18.80&18.41&17.09&\textbf{16.58}\\
&Shake-Shake (26 2x96d)&17.05&16.00&\textbf{14.28}&15.20\\
\textbf{Reduced CIFAR-10}&Wide-ResNet-28-10&18.84&16.50&14.13&\textbf{13.36}\\
&Shake-Shake (26 2x96d)&17.05&13.40&\textbf{10.04}&10.26\\
\hline
\end{tabular}}
\end{table}

\subsubsection{SVHN Results}\label{SVHN Results}
Based on the same experimental setup as CIFAR, we ran BO 8 times on Reduced SVHN which contains 1,000 examples to select the optimal policies for the final models training. The selected policies are shown in Table \ref{tab7}, and they will be later used on Reduced SVHN and SVHN.
\begin{table}[!htb]
\footnotesize
\centering
\caption{\label{tab7} The optimal policies found on Reduced SVHN. All the geometric-based operations are bolded.}
\centering
\begin{tabular}{lll}
\hline
&\textbf{Operation 1}&\textbf{Operation 2}\\
\hline
Sub-policy 1&\textbf{(ShearY, 0.7614, 1.9849)}&(Brightness, 0.0097, 8.1271)\\
Sub-policy 2&\textbf{(Invert, 0.7468, 3.1175)}&\textbf{(ShearY, 0.6576, 4.8592)}\\
Sub-policy 3&(Brightness, 0.9610, 2.3202)&(Equalize, 0.9848, 2.5281)\\
Sub-policy 4&(Contrast, 0.7195, 7.4252)&(Color, 0.3982, 3.3927)\\
Sub-policy 5&\textbf{(TranslateX, 0.3231, 2.1326)}&(Solarize, 0.9329, 7.0658)\\
Sub-policy 6&(Equalize, 0.8227, 3.3000)&\textbf{(Rotate, 0.7638, 3.0728)}\\
Sub-policy 7&(Equalize, 0.9919, 7.0251)&\textbf{(TranslateY, 0.6032, 4.9770)}\\
Sub-policy 8&(Solarize, 0.5896, 8.5091)&\textbf{(Invert, 0.6628, 6.2895)}\\
Sub-policy 9&\textbf{(Rotate, 1.0000, 7.4837)}&\textbf{(ShearY, 0.2407, 6.0313)}\\
Sub-policy 10&\textbf{(Invert, 0.8723, 7.2289)}&(Posterize, 0.3586, 7.1004)\\
Sub-policy 11&(Color, 0.0072, 4.7249)&\textbf{(Invert, 0.9959, 3.8010)}\\
Sub-policy 12&(Solarize, 0.2024, 3.0107)&(Color, 0.7671, 5.0435)\\
Sub-policy 13&\textbf{(Invert, 0.6668, 5.4347)}&\textbf{(Rotate, 0.4755, 4.1415)}\\
Sub-policy 14&(Contrast, 0.9712, 8.2252)&(Color, 0.8781, 8.2405)\\
Sub-policy 15&(Brightness, 0.4671, 8.6339)&(AutoContrast, 0.6229, 5.1071)\\
Sub-policy 16&\textbf{(ShearY, 0.9499, 3.2718)}&(Brightness, 0.4102, 3.7130)\\
Sub-policy 17&(Solarize, 0.7952, 7.3589)&\textbf{(ShearY, 0.5424, 3.7230)}\\
Sub-policy 18&(Contrast, 0.9796, 7.3553)&(Solarize, 0.0061, 2.4950)\\
Sub-policy 19&(Posterize, 0.9873, 2.2049)&\textbf{(Invert, 0.4173, 6.7201)}\\
Sub-policy 20&\textbf{(Invert, 0.6843, 1.2672)}&(ShearY, 0.6295, 2.2341)\\
Sub-policy 21&(Contrast, 0.9862, 3.8489)&(Equalize, 0.7513, 3.1612)\\
Sub-policy 22&(Color, 0.6640, 2.4163)&(Solarize, 0.3826, 8.3206)\\
Sub-policy 23&\textbf{(TranslateY, 0.0706, 6.1889)}&(Contrast, 0.0941, 0.0007)\\
Sub-policy 24&(Equalize, 0.3803, 4.7222)&\textbf{(TranslateY, 0.5262, 7.7046)}\\
\hline
\end{tabular}
\end{table}
As we can see from Table \ref{tab7}, the policies picked on SVHN are different from what are found on CIFAR-10. Geometry-based operations are mainstream here (20 operations in total and with high application probability), such as Invert that is hardly used on CIFAR-10, where it is always applied with a high probability on SVHN. This finding makes sense from the essential attributes of SVHN dataset: house number are often naturally sheared and tilted, so it is helpful to learn the invariance with DA techniques. Despite this, color-based operations still account for a large percentage about 58\%, which is slightly different from AutoAugment's \cite{cubuk2018autoaugment} 30\% conclusion. Then, we apply the selected policies on Reduced SVHN and SVHN datasets. Note that we do not use the Cutout operation in the baseline pre-processing on the Reduced SVHN. The summary of the results in this experiment are shown in Table \ref{tab8}.

Again, our policies selected at a very low computational cost achieve comparable results to AutoAugment \cite{cubuk2018autoaugment}.
\begin{table}[!htb]
\footnotesize
\centering
\caption{\label{tab8} Test set error rates (\%). Lower error rates are better. All the baseline and AutoAugment experiment results are reported in \cite{cubuk2018autoaugment}.}
\centering
\resizebox{\textwidth}{!}{
\begin{tabular}{llcccc}
\hline
\textbf{Dataset}&\textbf{Model}&\textbf{Baseline}&\textbf{Cutout}&\textbf{AutoAugment}&\textbf{BO-Aug}\\
\hline
\textbf{Reduced SVHN}&Wide-ResNet-28-10&13.21&32.5&\textbf{8.15}&8.19\\
&Shake-Shake (26 2x96d)&12.32&24.22&\textbf{5.92}&7.21\\
\textbf{SVHN}&Wide-ResNet-28-10&1.50&1.30&\textbf{1.07}&1.17\\
&Shake-Shake (26 2x96d)&1.40&1.20&\textbf{1.02}&1.15\\
\hline
\end{tabular}}
\end{table}

\subsection{Convergence Speed of BO-Aug}\label{Convergence Speed of BO}
\begin{figure}[!htb]
\centering
\includegraphics[width=13cm,height=8cm]{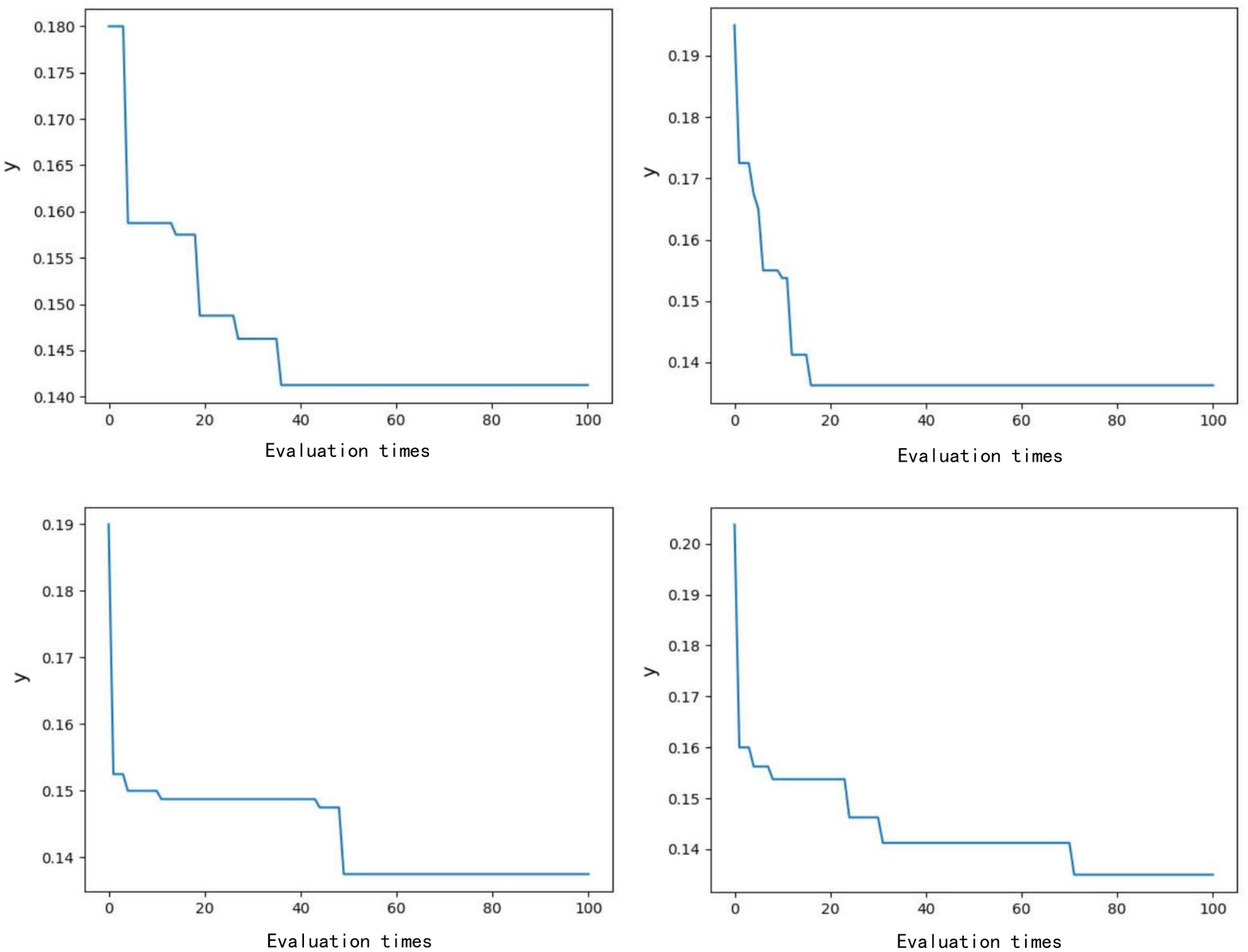}
\caption{Convergence curves of 4 single run on Reduced CIFAR-10 dataset. y-axis represents the classification error of the feedback model. (a) is the convergence curve for obtaining the optimal sub-policies 4, 5 and 6. (b) is the convergence curve for obtaining the optimal sub-policies 7, 8 and 9. (c) is the convergence curve for obtaining the optimal sub-policies 10, 11 and 12. (d) is the convergence curve for obtaining the optimal sub-policies 16, 17 and 18. The optimal sub-policies found on Reduced CIFAR-10 is shown in Table \ref{tab6}.}
\label{fig3}
\end{figure}

For each reduced dataset, we run BO 8 times independently to get the final optimal DA policies, and the number of iterations for a single run is set to 100. However, in fact, the optimal policy is usually obtained less than 100 real evaluations on each run of BO. Here we take the Reduced CIFAR-10 dataset as an example and give the convergence curves of 4 single run in Figure \ref{fig3}. The results on Reduced SVHN dataset is similar.

\subsection{Discussion}\label{Discussion}
Here we give the answers to the 3 questions raised at the beginning of this section.

\noindent\textbf{Effect of Data Augmentation}: Looking at all the experimental results, we can see that the improvement in accuracy due to BO-Aug is more significant on the reduced datasets compared to the full datasets, and BO-Aug on both reduced datasets results are comparable to state-of-the-art semi-supervised methods. This result is actually in line with our expectations, that is, the smaller the amount of data, the greater the effectiveness of data augmentation technique can play.

\noindent\textbf{Effectiveness and high efficiency of BO-Aug}: As can be seen from the experimental results, BO-Aug has a good performance on all data sets. It can always outperform the baseline methods or obtain comparable results. In addition, it shows high efficiency in policy searching. Training deep neural network models is a time consuming and computational resource consuming task. By way of illustration, for the feedback model WRN-40-2 we used in this paper, it takes more than 20 minutes for a real evaluation of the selected policy based on single GeForce GTX 1080 Ti GPU. As mentioned in \cite{cubuk2018autoaugment}, using the same evaluation model WRN-40-2, AutoAugment needs 15,000 real evaluations on one dataset to select optimal policies for final use. In contrast, using BO-Aug requires only 800 real evaluations, and the time is less than 6\% of AutoAugment. Particularly, the running time of BO-Aug itself is negligible compared to the real evaluation time of the feedback model.

\noindent\textbf{Transferability across datasets and architectures}: One thing that needs special attention is that, although we finally validated the effectiveness of BO-Aug on 5 datasets, we actually only used 2 datasets to find optimal policies. This means that the policies learned by a specific feedback model on a dataset are transferable between different model architectures and datasets. For instance, the policies found on Reduced CIFAR-10 by WRN-40-2 can lead to the improvement on all of the other model architectures trained on CIFAR-10 and even on CIFAR-100. This undoubtedly bring a light to routine model training work, the DA policies found by BO-Aug can hopefully help researchers improve the models¡¯ generalization performance on relevant image classification tasks.

\section{Conclusion and Future Work}\label{Conclusion and Future Work}
In this paper, we proposed BO-Aug which can automatically learn data augmentation policies for image classification tasks. The whole model consists of two components, namely the construction of search space and finding the optimal policies by means of Bayesian optimization. We aim to learn the optimal DA policies with the lowest possible computational cost. During the policies search process, BO-Aug only needs 800 real evaluations, which is far less than 15,000 times of the existing method \cite{cubuk2018autoaugment}. Experimental results show that our method can achieve state-of-the-art or near-advanced classification accuracy with different model architectures on several widely used image classification datasets.

There are a few of possible future directions of research for learning DA policies in the proposed model, for example, considering the policies to be learned as a form of rules, so that the model will eventually learn the knowledge about how to augment the data, that is, which DA operation should be used when the image satisfies certain conditions. More broadly, we are excited about how to extend the model proposed in this paper to other tasks such as natural language processing and speech recognition.

\section*{Acknowledgments}
\section*{References}

\bibliographystyle{elsarticle-num}
\bibliography{ref}

\begin{thebibliography}{10}
\expandafter\ifx\csname url\endcsname\relax
  \def\url#1{\texttt{#1}}\fi
\expandafter\ifx\csname urlprefix\endcsname\relax\def\urlprefix{URL }\fi
\expandafter\ifx\csname href\endcsname\relax
  \def\href#1#2{#2} \def\path#1{#1}\fi

\bibitem{krizhevsky2012imagenet}
A.~Krizhevsky, I.~Sutskever, G.~E. Hinton, Imagenet classification with deep
  convolutional neural networks, in: Advances in neural information processing
  systems, 2012, pp. 1097--1105.

\bibitem{he2015spatial}
K.~He, X.~Zhang, S.~Ren, J.~Sun, Spatial pyramid pooling in deep convolutional
  networks for visual recognition, IEEE transactions on pattern analysis and
  machine intelligence 37~(9) (2015) 1904--1916.

\bibitem{girshick2014rich}
R.~Girshick, J.~Donahue, T.~Darrell, J.~Malik, Rich feature hierarchies for
  accurate object detection and semantic segmentation, in: Proceedings of the
  IEEE conference on computer vision and pattern recognition, 2014, pp.
  580--587.

\bibitem{redmon2016you}
J.~Redmon, S.~Divvala, R.~Girshick, A.~Farhadi, You only look once: Unified,
  real-time object detection, in: Proceedings of the IEEE conference on
  computer vision and pattern recognition, 2016, pp. 779--788.

\bibitem{collobert2008unified}
R.~Collobert, J.~Weston, A unified architecture for natural language
  processing: Deep neural networks with multitask learning, in: Proceedings of
  the 25th international conference on Machine learning, ACM, 2008, pp.
  160--167.

\bibitem{collobert2011natural}
R.~Collobert, J.~Weston, L.~Bottou, M.~Karlen, K.~Kavukcuoglu, P.~Kuksa,
  Natural language processing (almost) from scratch, Journal of machine
  learning research 12~(Aug) (2011) 2493--2537.

\bibitem{hinton2012deep}
G.~Hinton, L.~Deng, D.~Yu, G.~Dahl, A.-r. Mohamed, N.~Jaitly, A.~Senior,
  V.~Vanhoucke, P.~Nguyen, B.~Kingsbury, et~al., Deep neural networks for
  acoustic modeling in speech recognition, IEEE Signal processing magazine 29.

\bibitem{graves2013speech}
A.~Graves, A.-r. Mohamed, G.~Hinton, Speech recognition with deep recurrent
  neural networks, in: 2013 IEEE international conference on acoustics, speech
  and signal processing, IEEE, 2013, pp. 6645--6649.

\bibitem{hinton2012improving}
G.~E. Hinton, N.~Srivastava, A.~Krizhevsky, I.~Sutskever, R.~R. Salakhutdinov,
  Improving neural networks by preventing co-adaptation of feature detectors,
  arXiv preprint arXiv:1207.0580.

\bibitem{srivastava2014dropout}
N.~Srivastava, G.~Hinton, A.~Krizhevsky, I.~Sutskever, R.~Salakhutdinov,
  Dropout: a simple way to prevent neural networks from overfitting, The
  Journal of Machine Learning Research 15~(1) (2014) 1929--1958.

\bibitem{bouthillier2015dropout}
X.~Bouthillier, K.~Konda, P.~Vincent, R.~Memisevic, Dropout as data
  augmentation, arXiv preprint arXiv:1506.08700.

\bibitem{ioffe2015batch}
S.~Ioffe, C.~Szegedy, Batch normalization: Accelerating deep network training
  by reducing internal covariate shift, arXiv preprint arXiv:1502.03167.

\bibitem{sennrich2015improving}
R.~Sennrich, B.~Haddow, A.~Birch, Improving neural machine translation models
  with monolingual data, arXiv preprint arXiv:1511.06709.

\bibitem{hoang2018iterative}
V.~C.~D. Hoang, P.~Koehn, G.~Haffari, T.~Cohn, Iterative back-translation for
  neural machine translation, in: Proceedings of the 2nd Workshop on Neural
  Machine Translation and Generation, 2018, pp. 18--24.

\bibitem{prabhumoye2018style}
S.~Prabhumoye, Y.~Tsvetkov, R.~Salakhutdinov, A.~W. Black, Style transfer
  through back-translation, arXiv preprint arXiv:1804.09000.

\bibitem{ko2015audio}
T.~Ko, V.~Peddinti, D.~Povey, S.~Khudanpur, Audio augmentation for speech
  recognition, in: Sixteenth Annual Conference of the International Speech
  Communication Association, 2015.

\bibitem{rebai2017improving}
I.~Rebai, Y.~BenAyed, W.~Mahdi, J.-P. Lorr{\'e}, Improving speech recognition
  using data augmentation and acoustic model fusion, Procedia Computer Science
  112 (2017) 316--322.

\bibitem{lecun2015deep}
Y.~LeCun, Y.~Bengio, G.~Hinton, Deep learning, nature 521~(7553) (2015) 436.

\bibitem{cirecsan2012multi}
D.~Cire{\c{s}}an, U.~Meier, J.~Schmidhuber, Multi-column deep neural networks
  for image classification, arXiv preprint arXiv:1202.2745.

\bibitem{tran2017bayesian}
T.~Tran, T.~Pham, G.~Carneiro, L.~Palmer, I.~Reid, A bayesian data augmentation
  approach for learning deep models, in: Advances in Neural Information
  Processing Systems, 2017, pp. 2797--2806.

\bibitem{lemley2017smart}
J.~Lemley, S.~Bazrafkan, P.~Corcoran, Smart augmentation learning an optimal
  data augmentation strategy, IEEE Access 5 (2017) 5858--5869.

\bibitem{perez2017effectiveness}
L.~Perez, J.~Wang, The effectiveness of data augmentation in image
  classification using deep learning, arXiv preprint arXiv:1712.04621.

\bibitem{antoniou2017data}
A.~Antoniou, A.~Storkey, H.~Edwards, Data augmentation generative adversarial
  networks, arXiv preprint arXiv:1711.04340.

\bibitem{zhu2017data}
X.~Zhu, Y.~Liu, Z.~Qin, J.~Li, Data augmentation in emotion classification
  using generative adversarial networks, arXiv preprint arXiv:1711.00648.

\bibitem{ratner2017learning}
A.~J. Ratner, H.~Ehrenberg, Z.~Hussain, J.~Dunnmon, C.~R{\'e}, Learning to
  compose domain-specific transformations for data augmentation, in: Advances
  in neural information processing systems, 2017, pp. 3236--3246.

\bibitem{cubuk2018autoaugment}
E.~D. Cubuk, B.~Zoph, D.~Mane, V.~Vasudevan, Q.~V. Le, Autoaugment: Learning
  augmentation policies from data, arXiv preprint arXiv:1805.09501.

\bibitem{shahriari2016taking}
B.~Shahriari, K.~Swersky, Z.~Wang, R.~P. Adams, N.~De~Freitas, Taking the human
  out of the loop: A review of bayesian optimization, Proceedings of the IEEE
  104~(1) (2016) 148--175.

\bibitem{jiaxu2017BOsurvey}
J.~Cui, B.~Yang, Survey on bayesian optimization methodology and applications,
  in: Journal of Software, 2017(in Chinese), pp. 176--198.

\bibitem{khajah2016designing}
M.~M. Khajah, B.~D. Roads, R.~V. Lindsey, Y.-E. Liu, M.~C. Mozer, Designing
  engaging games using bayesian optimization, in: Proceedings of the 2016 chi
  conference on human factors in computing systems, ACM, 2016, pp. 5571--5582.

\bibitem{frazier2016bayesian}
P.~I. Frazier, J.~Wang, Bayesian optimization for materials design, in:
  Information science for materials discovery and design, Springer, 2016, pp.
  45--75.

\bibitem{li2010contextual}
L.~Li, W.~Chu, J.~Langford, R.~E. Schapire, A contextual-bandit approach to
  personalized news article recommendation, in: Proceedings of the 19th
  international conference on World wide web, ACM, 2010, pp. 661--670.

\bibitem{vanchinathan2014explore}
H.~P. Vanchinathan, I.~Nikolic, F.~De~Bona, A.~Krause, Explore-exploit in top-n
  recommender systems via gaussian processes, in: Proceedings of the 8th ACM
  Conference on Recommender systems, ACM, 2014, pp. 225--232.

\bibitem{akrour2017local}
R.~Akrour, D.~Sorokin, J.~Peters, G.~Neumann, Local bayesian optimization of
  motor skills, in: Proceedings of the 34th International Conference on Machine
  Learning-Volume 70, JMLR. org, 2017, pp. 41--50.

\bibitem{torun2018global}
H.~M. Torun, M.~Swaminathan, A.~K. Davis, M.~L.~F. Bellaredj, A global bayesian
  optimization algorithm and its application to integrated system design, IEEE
  Transactions on Very Large Scale Integration (VLSI) Systems 26~(4) (2018)
  792--802.

\bibitem{bergstra2011algorithms}
J.~S. Bergstra, R.~Bardenet, Y.~Bengio, B.~K{\'e}gl, Algorithms for
  hyper-parameter optimization, in: Advances in neural information processing
  systems, 2011, pp. 2546--2554.

\bibitem{snoek2012practical}
J.~Snoek, H.~Larochelle, R.~P. Adams, Practical bayesian optimization of
  machine learning algorithms, in: Advances in neural information processing
  systems, 2012, pp. 2951--2959.

\bibitem{swersky2013multi}
K.~Swersky, J.~Snoek, R.~P. Adams, Multi-task bayesian optimization, in:
  Advances in neural information processing systems, 2013, pp. 2004--2012.

\bibitem{mahendran2012adaptive}
N.~Mahendran, Z.~Wang, F.~Hamze, N.~De~Freitas, Adaptive mcmc with bayesian
  optimization, in: Artificial Intelligence and Statistics, 2012, pp. 751--760.

\bibitem{netzer2011reading}
Y.~Netzer, T.~Wang, A.~Coates, A.~Bissacco, B.~Wu, A.~Y. Ng, Reading digits in
  natural images with unsupervised feature learning.

\bibitem{zagoruyko2016wide}
S.~Zagoruyko, N.~Komodakis, Wide residual networks, arXiv preprint
  arXiv:1605.07146.

\bibitem{williams2006gaussian}
C.~K. Williams, C.~E. Rasmussen, Gaussian processes for machine learning,
  Vol.~2, MIT Press Cambridge, MA, 2006.

\bibitem{carr2016basc}
S.~Carr, R.~Garnett, C.~Lo, Basc: applying bayesian optimization to the search
  for global minima on potential energy surfaces, in: International Conference
  on Machine Learning, 2016, pp. 898--907.

\bibitem{mockus1978application}
J.~Mockus, V.~Tiesis, A.~Zilinskas, The application of bayesian methods for
  seeking the extremum, Towards global optimization 2~(117-129) (1978) 2.

\bibitem{hansen2001completely}
N.~Hansen, A.~Ostermeier, Completely derandomized self-adaptation in evolution
  strategies, Evolutionary computation 9~(2) (2001) 159--195.

\bibitem{krizhevsky2009learning}
A.~Krizhevsky, G.~Hinton, Learning multiple layers of features from tiny
  images, Tech. rep., Citeseer (2009).

\bibitem{gastaldi2017shake}
X.~Gastaldi, Shake-shake regularization, arXiv preprint arXiv:1705.07485.

\bibitem{devries2017improved}
T.~DeVries, G.~W. Taylor, Improved regularization of convolutional neural
  networks with cutout, arXiv preprint arXiv:1708.04552.

\bibitem{miyato2018virtual}
T.~Miyato, S.-i. Maeda, S.~Ishii, M.~Koyama, Virtual adversarial training: a
  regularization method for supervised and semi-supervised learning, IEEE
  transactions on pattern analysis and machine intelligence.

\end{thebibliography}

\end{document}